\documentclass[10pt, a4paper]{article}

\usepackage{styles/arxiv}         
\usepackage[utf8]{inputenc}

\usepackage[english]{babel}
\usepackage[T1]{fontenc}    
\usepackage{hyperref}       
\usepackage{url}            
\usepackage{booktabs}       
\usepackage{amsfonts}       
\usepackage{microtype}      
\usepackage{amsmath}
\usepackage{amssymb}        
\usepackage{amsthm}         
\usepackage{dsfont}         
\usepackage{float}
\usepackage{graphicx}
\usepackage{stmaryrd}       
\usepackage{tikz}
\usepackage{csquotes}
\usepackage{tabularx}
\usepackage{multirow}

\usepackage{pgfplots}
\usepgfplotslibrary{colormaps}

\usepackage{todonotes}
\usepackage{caption}
\usepackage{subcaption}     

\usepackage{algorithm}
\usepackage{algpseudocode}

\usepackage{lipsum}

\usepackage{natbib}
\pgfplotsset{compat = newest}
\pgfplotsset{
    colormap={uni}{rgb255(0cm)= (0, 170, 220); rgb255(1cm)= (255, 20, 11)}
}

\definecolor{BlueUnilu}{RGB}{0, 170, 220}
\definecolor{RedUnilu}{RGB}{255, 20, 11}

\newcommand\cmp[3]{((#1^x)/((#3) * (x!)^#2))}

\title{Prediction of Handball Matches with Statistically Enhanced Learning via Estimated Team Strengths}

\date{} 					

\author{
 Florian Felice\footnote[1]{1}\\
 Department of Mathematics\\
 University of Luxembourg\\
 \href{mailto:florian.felice@uni.lu}{\texttt{florian.felice@uni.lu}}\\
 \And
 Christophe Ley\\
 Department of Mathematics\\
 University of Luxembourg\\
 \href{mailto:christophe.ley@uni.lu}{\texttt{christophe.ley@uni.lu}}\\
}

\begin{document}

\pgfplotsset{
    colormap={greenred}{rgb255(0cm)=(0, 220, 0); rgb255(0.75cm)=(51, 102, 0); rgb255(0.85cm)=(0, 0, 0); rgb255(0.95cm)=(204, 0, 0); rgb255(1cm)=(255, 20, 11)}
}

\maketitle

\renewcommand{\thefootnote}{*}
\footnotetext{Work not related to Amazon}
\renewcommand{\thefootnote}{\arabic{footnote}}

\begin{abstract}
	We propose a Statistically Enhanced Learning (aka. SEL) model to predict handball games.
	Our Machine Learning model augmented with SEL features outperforms state-of-the-art models with an accuracy beyond 80\%.
	In this work, we show how we construct the data set to train Machine Learning models on past female club matches.
	We then compare different models and evaluate them to assess their performance capabilities.
	Finally, explainability methods allow us to change the scope of our tool from a purely predictive solution to a highly insightful analytical tool.
	This can become a valuable asset for handball teams' coaches providing valuable statistical and predictive insights to prepare future competitions.
\end{abstract}


\section{Introduction}\label{sec:intro}

Handball is a popular sport in Europe with growing interest in Northern Africa and South America.
As a fast-paced sport, it is gaining interest in the population and in the scientific literature though predictive models are rarely discussed.

In this paper, we propose a Statistically Enhanced Learning (aka. SEL) model to predict handball matches.
SEL is a general approach \citep{felice_statistically_2023} that aims to formalize the feature extraction step from feature engineering when preparing data for a Machine Learning (ML) model.
It is shown that generating SEL features can lead to a considerable boost of predictive performance while providing meaningful statistical covariates that can be interpretable.
Unlike other sports, handball does not benefit from a large literature, and in particular  predictive algorithms are scarce.
In Section~\ref{sec:data_models}, we will describe the construction of our training data set based on publicly available data.
We will explore different ML models and show how the SEL methodology helps improve their predictive performance.
Next, we will present the results of the trained models in Section~\ref{sec:results} and show how we can extract informative sports insights from a ML model via an explainable ML framework.
Section~\ref{sec:discussion} shows that  this paper not only yields a good performing predictive model, but that it can also serve as user-friendly tool to team coaches in view of preparing upcoming games and competitions.
Indeed, our proposal includes a highly performing ML model with, on top, explainability modules that can allow teams to identify important factors impacting the games' results.
Finally, we conclude in Section~\ref{sec:conclusion}.

\subsection{The history of handball}

With its primitive form going back to the ancient Greece, modern handball was considered to be created by German sports teachers (outdoors with 11-aside players) around 1890 while Scandinavian countries (Denmark and Sweden) introduced a version with 7-aside players around the same period \citep{hahn_fascination_2013}.
Its original Danish name ``Haandbold'' was first called in 1898 and the first official competition was organized in 1917 when the term ``handball'' was also officially used for the first time.
It became an Olympic discipline at the 1972 Olympics in Munich for men and at the 1976 Olympics in Montreal for women \citep{olympics_history_2023}.

\subsection{Literature review and related work}

Sports predictions is an active field of research mostly focusing on sports such as football and basketball due to a larger amount of data publicly available.
To predict the outcome of a match, several algorithms are considered to model sports matches such as linear regressions \citep{miljkovic_use_2010, rodriguez-ruiz_study_2011}, support vector machines \citep{cai_hybrid_2019}, random forests \citep{groll_hybrid_2019}, XGBoost \citep{lampis_predictions_2023} or neural networks \citep{mccabe_artificial_2008, huang_neural_2010}.
In the field of football predictions, \cite{groll_hybrid_2019} use a Random Forest to predict the outcome of football matches.
They augment their data set by adding a feature corresponding to the strength of a team, in the principle of Statistically Enhanced Learning \citep{felice_statistically_2023}.
This value is obtained by modelling scored goals with a bivariate Poisson distribution and assuming that the form of the estimated parameter is $\log(\lambda_i) = \beta_0 + (r_i - r_j) + h \cdot \mathds{1}(\text{team } i \text{ playing at home})$.
The parameter $\beta_0$ corresponds to a common intercept and $h$ is the effect of playing at home.
The values of interest are then $r_i$ and $r_j$ as the strength parameters of the home team $i$ and away team $j$, estimated via Maximum Likelihood Estimation.
These new values will become a new feature in the data set to enhance the future learning algorithm.

Only scarce literature covers the field of handball analytics \citep{saavedra_handball_2018}.
Most of the existing research works are medical analyses looking at body fatigue and injury \citep{akyuz_skeletal_2019, seil_sports_1998, camacho-cardenosa_anthropometric_2018}
with a particular interest on young players \citep{madsen_activity_2019,grabara_posture_2018,fonseca_relative_2019}.
\cite{wagner_individual_2014} propose a review of performance for handball players and teams, highlighting the importance of factors such as experience level, age or playing positions.
\cite{pic_performance_2018} showed that, the impact of the home effect can play a role in critical moments of a game.
In particular, when the score indicates a draw, the home team may finally win the game.
As highlighted by the author, this advantage should be taken with care as it can either come from the effect of playing at home (with more supporters cheering) or from the fatigue of the away team from the travel.
Indeed, unlike sports such as football mostly travelling by plane or train, most handball clubs travel by bus which can have an important impact on the players fatigue during the competition.
Therefore, the difference in distance traveled between both teams can explain the level of players' freshness.

With a focus on prediction of handball outcomes, \cite{groll_prediction_2020} compared different regression approaches to model international handball games.
Given the level of under-dispersion (when the variance is lower than the mean, i.e. $\mathds{V}(X) < \mathds{E}(X)$) they discarded the regular Poisson distribution and opted for a Gaussian distribution with low variance.
In a similar spirit, \cite{felice_ranking_2023} proposed to model the number of goals scored by a team with a Conway-Maxwell-Poisson distribution \citep{sellers_conway-maxwell-poisson_2022} and derive a strength parameter from the parameters of the fitted distribution.

In this work, we propose a machine learning approach to model handball matches.
Our model is also follows the framework of Statistically Enhanced Learning given that we add statistically estimated features to represent the strength of the teams.
In the following, we describe the data collected and features created in Section~\ref{sec:data_models}.
In Section~\ref{sec:results}, we present the results of trained machine learning models and compare their performance with the additional SEL generated features.
We discuss potential extensions and the use of our proposed approach in Section~\ref{sec:discussion} and we conclude in Section~\ref{sec:conclusion}.

\section{Materials and methods}\label{sec:data_models}

In this section we present the data used to train our machine learning models with the associated features.

\subsection{Data set}

Our data set consists of two data sources that include historical games and team squads information.
We extract information of past games using the \href{https://rapidapi.com/tipsters/api/sportscore1}{SportScore} API from the service \href{https://rapidapi.com}{RapidAPI}.
The extracted data include information such as game location, time, competition and score.
The second source used to complete the data set contains information about teams and players.
The data are extracted from the website \href{https://www.handball-base.com/}{\texttt{www.handball-base.com}} and will help us generate teams and players specific variables.

\subsubsection{Target response}

Our objective being to predict the number of goals scored by each team by the end of a match, our modelling exercise is a two-target regression problem.
We then predict the final score for home and away teams.
We can also use the score to determine for a team (e.g., the home team) whether the game was won, lost or ended in a draw, and train a classification model on this basis.

\subsubsection{Features}\label{sec:features}

Our data set is composed of features which bring different levels of information about the two competing teams.
The exhaustive list of features with abbreviations is available in Appendix~\ref{apx:features}.

\paragraph{Game information}
These features aim to carry information about the game and its importance.
It can help encapsulate information such as potential stress for players and their state of mind (for instance, how seriously they may take a game).
\begin{itemize}
	\item \texttt{Day of week}: encoded day of the week for the start time of the game.
	\item \texttt{Hour}: hour of the start time of the game. We can expect that games starting early in the day (e.g., morning) can be less important or players may lack time for preparation.
	\item \texttt{Importance}: carries the importance of the competition from the lowest (friendly games with value 3) to the highest importance (Champions League with value 1).
	\item \texttt{Days until final}: counts the number of days until the competition's final or end of season. Combined with the variable \texttt{Importance}, this should account for the intensity of the game (last day of competition for the championship -- i.e. final -- may potentially be more important than the first day of the Champions League).
\end{itemize}

\paragraph{Teams' structure}
We also consider features allowing us to capture information from the team's physical abilities and experience, and we incorporate them as differences between the home and away teams.
There is one feature for each attribute (height, weight, age) and position on the field (wings, back players, line players/pivots and goalkeepers).
This set of features can also be seen as SEL variables (as defined in \cite{felice_statistically_2023}) but we will refer to them as classical variables.

\begin{itemize}
	\item \texttt{Height}: difference of the average height of players per position between home and away teams. This aims to measure the difference in physical characteristics between players (e.g., taller back players for one team may result in an advantage both in attack and in defense).
	\item \texttt{Weight}: difference of the average weight of players per position between home and away teams. Similar to \texttt{Height}, this aims to capture the differences  in physical abilities between players.
	\item \texttt{Age}: difference of the average age of players per position between home and away teams. This aims to capture the difference in experience/maturity between teams. The effect of such feature is expected to have a concave shape, with players becoming more performing when gaining experience (and still being young) up until their career peak after which they will start dropping in performance.
\end{itemize}

Other features give us information about the team's structure such as the distance to travel or the team's composition.

\begin{itemize}
	\item \texttt{Travel distance}: distance in kilometers (as the crow flies) to travel for the away team between the club's location and the address of the home team. This aims to capture the potential fatigue caused by the travelling distance.
	\item \texttt{Nationalities}: ratio between the total count of players' nationalities in a team and the total amount of players. This aims to capture the affinity between players as well as  potential language barriers.
	\item \texttt{International}: share of players selected in their national team.
\end{itemize}

\paragraph{Team's strength}

We finally add features that correspond to the teams' strength as described in more details in Section~\ref{sec:sel_feat}.
These variables are the aforementioned SEL features.

\begin{itemize}
	\item \texttt{Attack strength}: estimated strength in attack via SEL for home and away teams.
	\item \texttt{Defense strength}: estimated strength in defense via SEL for home and away teams.
\end{itemize}

\subsection{Estimating team strengths}\label{sec:sel_feat}

\begin{figure}[!ht]
	\centering
	\begin{tikzpicture}
    \begin{axis}[
    height=7cm,
    width=10cm,
    xmin=10,
    xmax=50,
    xlabel = Number of goals,
    ylabel = Frequency,
    legend style={nodes={scale=0.7, transform shape}},
    x tick label style={draw=none},
    y tick label style={font=\footnotesize, draw=none},
    ]

   \addplot[
       black,
       fill=lightgray,
       hist=density,
       hist/bins=20,
   ] table[y=y] {figures/data/dist.csv};

   \addplot[domain={10:50}, samples=40, dashed] {\cmp{250.45803272774776}{1.6390174920550273}{4193921903904605339648/3}};
   \addlegendentry{Empirical}
   \addlegendentry{$CMP(286.46, 1.64)$}

   \end{axis}
\end{tikzpicture}
	\caption{Histogram of goals scored by Metz Handball during the season 2021/2022 versus fitted Conway-Maxwell-Poisson (CMP) distribution.}
	\label{fig:distri}
\end{figure}
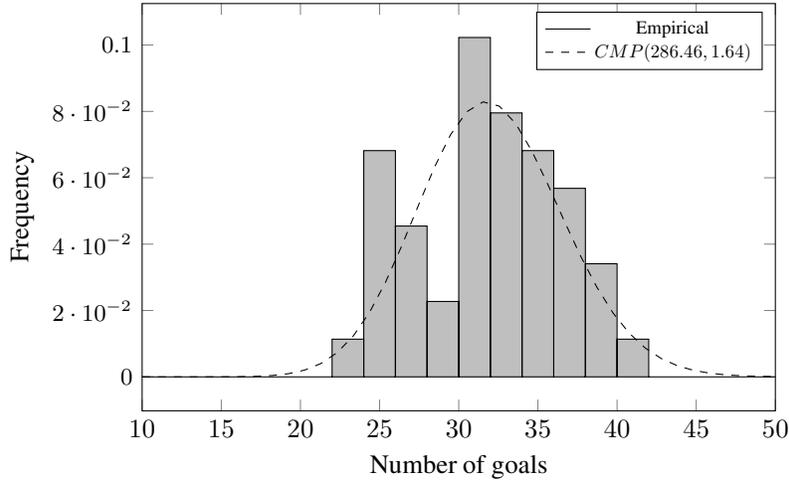

The strength of a team is an undeniably important factor of a handball match but it is not directly measurable and only remains an abstract concept.
We can palliate this shortcoming by devising a statistical model that incorporates parameters which are meant to represent the attacking and defensive strengths of each team, and then estimate these parameters. 
To this end, we consider the recent history (from the ongoing season) of each team's matches and fit the distribution of scored goals with an appropriate probability law.
\cite{felice_ranking_2023} explained that the Conway-Maxwell-Poisson distribution~\citep{sellers_conway-maxwell-poisson_2022} is a very good choice for this purpose as it not only satisfies the discrete nature of goal counts but also handles the problem of over- and under-dispersion one may have to deal with.
Hence it is a better choice than the often used Normal (not discrete), Poisson (assumes equi-dispersion) and Negative Binomial (cannot handle under-dispersion) distributions, for instance.
An illustration of the fitted distribution on historical data is provided in Figure~\ref{fig:distri} using the history of Metz Handball games over the course of the season 2021/2022.

The Conway-Maxwell-Poisson distribution possesses two parameters, $\lambda > 0$ and $\nu \geq 0$ (note that $\nu = 0$ implies that $\lambda \in (0,1)$).
The parameter $\lambda$ can be assimilated with the empirical mean (depending on the values of $\nu$).
For instance, when $\nu=1$ we retrieve the Poisson distribution for which $\lambda$ corresponds to both the mean and variance. 
The parameter $\nu$ corresponds to the level of dispersion.

Based on the nature of the two parameters, \cite{felice_ranking_2023} proposes  to penalize irregularities of the team, and consequently defines the strengths of a team as

\begin{equation}\label{eq:strength}
	s_a = \dfrac{\log(\lambda_a)}{\nu_a} \text{\quad and \quad} s_d = \dfrac{\nu_d}{\log(\lambda_d)},
\end{equation}

where $\lambda_a,\nu_a$ stand for the attacking parameters of a team, and $\lambda_d,\nu_d$ for the defensive parameters of the same team.
A strong attack demonstrates a high average number of scored goals ($\lambda_a$) and a balanced dispersion ($\nu_a$, stable with occasional spikes).
On the other hand, a strong defense corresponds to a low average of conceded goals ($\lambda_d$) and a stable defense over matches (low variance translating into under-dispersion with high value for $\nu_d$).

We thus model historical matches with the Conway-Maxwell-Poisson distribution.
We estimate, for each team, the defense parameters ($\lambda_d$ and $\nu_d$) as well as the attack parameters ($\lambda_a$ and $\nu_a$).
This consequently gives us the estimated strength parameters $s_a$ and $s_d$ that will constitute the SEL variables as presented in Section~\ref{sec:features}.

\subsection{Prediction models}\label{sec:models}

To model the outcome of a handball match, we consider both classification and regression models to either predict the winner of the game or the scores of the competing teams.
The results of the experimented models for classification and regression are discussed in Section~\ref{sec:results}.

\subsubsection{Classification model}

To predict the outcome of a match (as win, draw or loss), we train different ML classification algorithms.
A first approach is based on Random Forests \citep{breiman_random_2001}.
Another model is based on the popular XGBoost algorithm \citep{chen_xgboost_2016}.
We also use an improved version of the boosting model, CatBoost \citep{prokhorenkova_catboost_2018}, which is specialized in handling categorical data.
Finally, we train a Multi-Layered Perceptron \citep{rosenblatt_perceptron_1958}.

\subsubsection{Regression model}

As insightful alternative, we also consider regression models to predict the score of each team during a match.
To that end, we use multi-target variants of the aforementioned models which will predict the final score of the home and away teams.
We note that, by nature, the Random Forest model does not support multi-target regressions and can only predict one outcome.
We thus use Python's modules to implement these models\footnote{Using the module \href{https://scikit-learn.org/stable/modules/generated/sklearn.multioutput.MultiOutputRegressor.html}{\texttt{MultiOutputRegressor}} from the \href{https://scikit-learn.org/}{scikit-learn} library}.
The module is a wrapper class of the model that, under the hood, simply trains two models in parallel: one for the home and one for the away team.
The implementation for XGBoost does not support multi-target either.
Therefore, we also use the same Python module alternative to achieve our goal.
The other models, CatBoost and Multi-Layered Perceptron, can handle multiple outcome predictions by nature.

\subsection{Performance metrics}\label{sec:metrics}

To evaluate the performance of our models, we first define the metrics we use for classification and regression exercises.

\subsubsection{Metrics for classification models}
To measure performance of our classification models, we use three common metrics in the field of sports predictions: accuracy, weighted $F_1$-score and Brier score.

The $F_1$-score is the mean between the precision and recall but can also be written as the ratio between the true positives ($TP$) from the confusion matrix with the false positives ($FP$) and false negatives ($FN$) such as

\begin{equation}
    F_1 = 2 \cdot \dfrac{2 TP}{2 TP + FP + FN}.
\end{equation}

It is a popular classification metric which puts the same weight between precision (share of predictions being correct) and recall (share of positive items being predicted).

The Brier score \citep{brier_verification_1950} is a metric that is used to calculate the distance, in terms of probability, to the actual outcome.
It is defined as

\begin{equation}
    BS = \dfrac{1}{n} \sum_{i=1}^{n} (y_i - f(x_i))^{2}.
\end{equation}

It computes the squared difference between the predicted probability and the actual outcome (e.g. 0 if the outcome is negative and 1 otherwise).
This metric is particularly popular for sports predictions in the context of classification.


\subsubsection{Metrics for regression models}

To assess the quality of our regression models, we use the Root Mean Squared Error (RMSE) and the Mean Absolute Percentage Error (MAPE).

The Root Mean Squared Error is defined as
\begin{equation}
    RMSE = \sqrt{\dfrac{1}{n} \sum_{i=1}^{n} (y_i - f(x_i))^{2}}.
\end{equation}

It computes the average deviation of our model's predictions from the actual scores.
Using the square as the power for the difference helps penalize more when the model gives extremely incoherent predictions.
On the other hand, when predictions are close to the actual outcome, the penalty is lower.

The Mean Absolute Percentage Error is defined as
\begin{equation}
    MAPE = \dfrac{1}{n} \sum_{i=1}^{n} \left|\dfrac{y_i - f(x_i)}{y_i}\right|.
\end{equation}

It computes the relative difference between the actual and predicted values and uses the absolute value that puts the same weight to large or small deviations.

\section{Results}\label{sec:results}

To evaluate the performance of our distinct approaches, we train the different models on several years of female club matches.
Our training set spans from September 2019 until April 2023 (representing 3,260 games) and leaves matches from April to June 2023 (250 games) as the test set.

In both the classification and regression settings, we train the four different models presented in Section~\ref{sec:models} and compare the scenarios with and without the SEL features introduced in Section~\ref{sec:sel_feat}.
The performance metrics are summarized in Section~\ref{sec:models_perf}.
After that comparison, we further investigate the best performing model with explainability frameworks for global and local explanations  (Section~\ref{sec:model_explain}).

\subsection{Model performances}\label{sec:models_perf}

In the first case of match classification, we train our four models and report the classification metrics evaluated on the test set in Table~\ref{tab:model_class_perf}.

\begin{table}[H] 
\caption{Classification models performance comparison based on accuracy, weighted $F_1$-score and Brier score.
Each model is considered once only based on classical covariates and once with the additional SEL variables. \label{tab:model_class_perf}}
\newcolumntype{C}{>{\centering\arraybackslash}X}
\begin{tabularx}{\textwidth}{CCCCC}
\toprule
\textbf{Model}					& \textbf{Features}	& \textbf{Accuracy}	& \textbf{$F_1$-score}	& \textbf{Brier-score}\\
\midrule
\multirow{2}{*}{Random Forest}	& Classical			& 60.11\%			& 57.64\%				& 0.4837\\
								& Classical + SEL				& \textbf{81.32\%}	& \textbf{79.15\%}		& \textbf{0.3145}\\
\midrule
\multirow{2}{*}{XGBoost}		& Classical			& 57.51\%			& 55.87\%				& 0.6189\\
								& Classical + SEL				& 73.57\%			& 71.06\%				& 0.5784\\
\midrule
\multirow{2}{*}{CatBoost}		& Classical			& 58.29\%			& 54.82\%				& 0.5181\\
								& Classical + SEL				& 79.57\%			& 77.04\%				& 0.3517\\
\midrule
\multirow{2}{*}{Neural Net}		& Classical			& 54.18\%			& 52.75\%				& 0.5371\\
								& Classical + SEL				& 68.08\%			& 66.67\%				& 0.4479\\
\bottomrule
\end{tabularx}
\end{table}

We can observe in Table~\ref{tab:model_class_perf} that the Random Forest with SEL features performs the best.
Furthermore, we can observe that adding SEL features to our models is always beneficial and, with no exception, strongly helps improve our metrics.
The performance improvement is particularly {remarkable} for the Random Forest model which has the highest gap between two scenarios.
Although its performance with classical covariates already achieves 60.11\%, the SEL features boost the performance to reach 81.32\% by adding the estimated strengths of the opposing teams.

Turning our attention to regression settings, we train multi-target regression models with and without SEL features and report the resulting performance metrics in Table~\ref{tab:model_reg_perf}.
Our metrics of interest here are the Root Mean Squared Error and Mean Absolute Percentage Error for both home and away teams.

\begin{table}[H]
	\caption{Regression models performance comparison based on Root Mean Squared Error and Mean Absolute Percentage Error. Each model is considered once only based on classical covariates and once with the additional SEL variables. Note that we separate the predictions for home and away teams.\label{tab:model_reg_perf}}
	\newcolumntype{C}{>{\centering\arraybackslash}X}
	\begin{tabularx}{\textwidth}{CCCCCC}
	\toprule
	 & & \multicolumn{2}{c}{\textbf{Home}} & \multicolumn{2}{c}{\textbf{Away}} \\
	\textbf{Model}					& \textbf{Features}	& \textbf{RMSE}	& \textbf{MAPE} & \textbf{RMSE}	& \textbf{MAPE}\\
	\midrule
	\multirow{2}{*}{Random Forest}	& Classical			& 5.05			& 14.91\% 		& 4.85			& 15.34\%\\
									& Classical + SEL				& 3.96			& 11.50\% 		& 3.79 			& 12.28\%\\
	\midrule
	\multirow{2}{*}{XGBoost}		& Classical			& 5.87			& 15.40\% 		& 5.16 			& 14.08\%\\
									& Classical + SEL				& 4.24			& 11.63\% 		& 4.09 			& \textbf{11.45\%}\\
	\midrule
	\multirow{2}{*}{CatBoost}		& Classical			& 5.13			& 14.86\% 		& 4.78 			& 15.17\%\\
									& Classical + SEL				& \textbf{3.79}			& \textbf{10.94\%} 		& \textbf{3.73} 			& 12.06\%\\
	\midrule
	\multirow{2}{*}{Neural Net}		& Classical			& 5.29			& 15.76\% 		& 5.07 			& 16.14\%\\
									& Classical + SEL				& 5.30			& 15.31\% 		& 4.96 			& 15.30\%\\
	\bottomrule
	\end{tabularx}
\end{table}

We can see from Table~\ref{tab:model_reg_perf} that, although the Random Forest with SEL achieves good performance levels, the CatBoost model can predict match scores with the least error.
Similar to our first classification use case, adding SEL features greatly benefits to all trained models.
This outcome suggests that our best model can accurately predict the outcome of a female handball game with an error of 3.8 goals for the home and away teams.
As a comparison, state-of-the-art predictive models for football can achieve a prediction error of 1.194 goals \citep{groll_hybrid_2019}.
Considering that the average number of goals during a game is 1.5 goals \citep{zebari_predicting_2021}, this corresponds to a 80\% error.
In our case for handball, the RMSE of our proposed model on our test set is 3.8 for 27.9 goals scored on average during a match.
This then corresponds to an error of 11\%.
This therefore highlights the reliability of our predictive models strengthened by the supplementary information carried by the SEL covariates.

\subsection{Model explainability}\label{sec:model_explain}

Explaining a model's outcome is crucial to trust its predictions and take actions from the generated explanations.
In this section, we explore the important features of the selected CatBoost model from Section~\ref{sec:models_perf} and show how the extracted SEL features are used by our model.
Furthermore, we show how explaining predictions can be used to drive actions that can help team coaches prepare an upcoming match.

Many ML models such as Random Forests or Neural Networks are considered to be black boxes: they are excellent in terms of prediction accuracy, but one cannot understand the factors that lead to a given prediction.
Therefore, explainability (aka model transparency) is an important capability any ML model should have.
Its importance will become even stronger with forthcoming regulations on Artificial Intelligence \citep{hamon_robustness_2020, sovrano_metrics_2022}.
We distinguish global explanations, which focus on analyzing the overall behavior of the model (importance of features), to local explanations, used to explain predictions of specific observations.
The model explainability literature covers a wide range of techniques from model specific approaches benefiting from the properties of a certain model \citep{du_techniques_2019} to model agnostic approaches.
Model agnostic solutions span from surrogate models such as LIME \citep{ribeiro_why_2016}, which aim to locally approximate the black box ML model with a simpler and interpretable model, to game theoretic based approaches.

For our setting, we use game-theoretic based approaches with the SHAP framework \citep{lundberg_unified_2017} to generate explanations.
In particular, given the structure of our model, we use the TreeSHAP implementation \citep{lundberg_local_2020} which uses the tree structure of the model to perform more efficient and exact calculations of Shapley values.

We note that, for the rest of this section, we focus on explainability of the regression model but the implementations and conclusions also perfectly apply to the classification model.

\subsubsection{Feature importance from global explanation}\label{sec:model_explain_feat_import}

\begin{figure}[!ht]
	\centering
	\resizebox{0.8\linewidth}{!}{
    \begin{tikzpicture}
\pgfplotstableread[col sep = comma]{figures/data/HandFeatImport.csv}\tabledm
\def\xlistmacro{}
\def\xliststring{}
\pgfplotstableforeachcolumnelement{index}\of\tabledm\as\entry{%
\xifinlist{\entry}{\xlistmacro}{}{
        \listxadd{\xlistmacro}{\entry}
        \edef\xliststring{\xliststring\entry,}
    }
}
        \begin{axis}[
            x tick label style={font=\footnotesize, draw=none, anchor=north},
            y tick label style={font=\footnotesize, draw=none, anchor=east},
            label style={font=\footnotesize, draw=none},
            width=0.8\linewidth, 
            height=0.8\linewidth,
            xbar, 
            bar width=5pt,
            xlabel = Importance,
            axis x line=bottom,
            axis y line=left,
            enlarge y limits=0.01, 
            symbolic y coords/.expand once={\xliststring},
            ytick = data,
         ]
            \addplot[BlueUnilu, fill=BlueUnilu] table [y=index, x=importance, col sep=comma] {figures/data/HandFeatImport.csv};
        \end{axis}
\end{tikzpicture}
    }
	\caption{Feature importance plot using TreeSHAP for predicting home team's goals}
	\label{fig:feat_imp}
\end{figure}
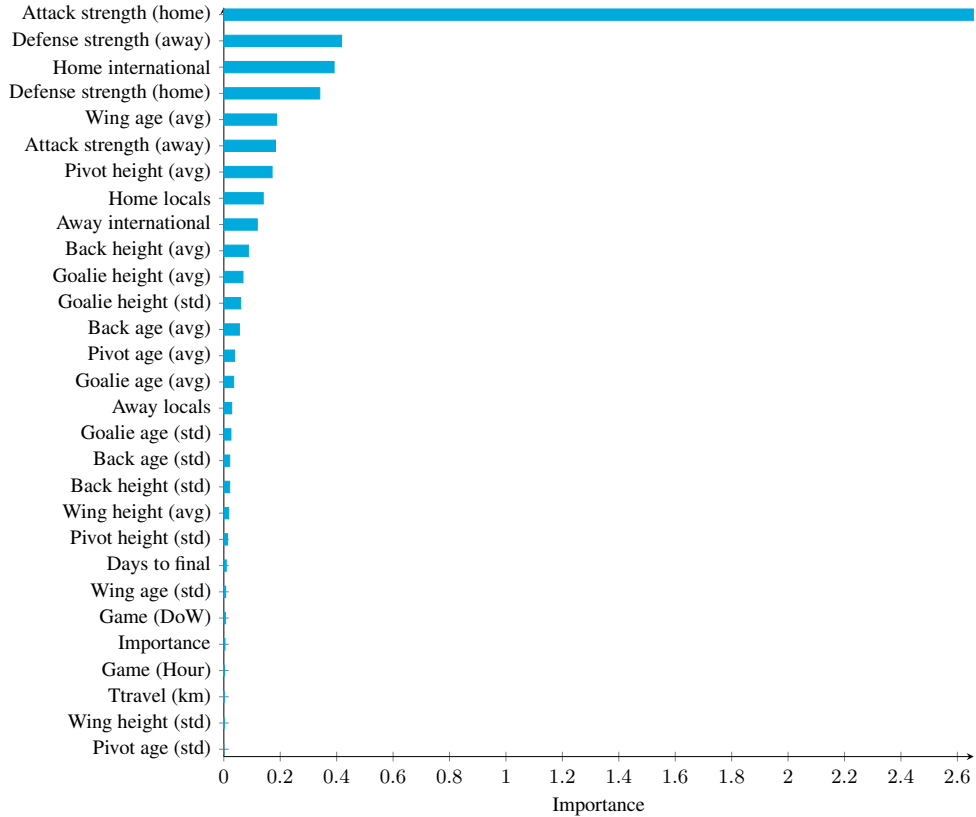

Analyzing feature importance can help understand the behavior of our model and identify influential variables.
Computing the Shapley values for each feature of the model, we can observe which are the ones being impactful to the predictions.
In other words, Figure~\ref{fig:feat_imp} tells us which features of our test set contribute the most to the predicted outcome.

We can observe that the SEL features for teams' strengths are considered as very important for the model.
We notice that the attack strength of the home team is the most important to predict home goals, followed by the defense strength from the away team.
This is perfectly logical, and again underlines the impact of SEL features.
To predict the score of the away team, we also observe that the most important feature is the attack strength of the away team.
It is then followed by the defense strength of the home team.
This is in line with conclusions from Section~\ref{sec:models_perf}.
We observed that, by adding these features to our model, the performance considerably improves.

\subsubsection{Understand match predictions from local explanations}\label{sec:model_explain_local}

Analyzing predictions by means of local explainability frameworks can help anticipate events during an upcoming match.
In a similar fashion as in the previous section, we use TreeSHAP \citep{lundberg_local_2020} to locally explain predictions.

\begin{figure}[!ht]
	\centering
	\resizebox{\linewidth}{!}{\begin{tikzpicture}
    \begin{axis}[
        colorbar, colormap name=greenred,
        scatter,
        scatter src=x,
        only marks,
        clip mode=individual,
        scatter/@pre marker code/.append code={
                \pgfkeys{/pgf/fpu=true,/pgf/fpu/output format=fixed}
                \pgfmathsetmacro\negheight{-\pgfplotspointmeta}         
                \fill [draw=black] 
                (axis direction cs:0,0.3) rectangle (axis direction cs:\negheight,-0.3);
                \pgfplotsset{mark=none}
            },
        legend columns=-1,
        ytick={0,...,28},
        yticklabels={Defense strength (away),Away international,Back age (avg),Goalie age (avg),Away locals,Goalie age (std),Back age (std),Importance,Wing height (std),Pivot age (std),Travel (km),Game (Hour),Game (DoW),Wing age (std),Days to final,Pivot height (std),Wing height (avg),Back height (std),Pivot age (avg),Goalie height (std),Goalie height (avg),Back height (avg),Home locals,Pivot height (avg),Attack strength (away),Wing age (avg),Defense strength (home),Home international,Attack strength (home))},
        x tick label style={font=\footnotesize, draw=none, anchor=north},
        y tick label style={font=\footnotesize, draw=none, anchor=east},
        label style={font=\footnotesize, draw=none},
        height=0.65\linewidth,
        xbar, 
        bar width=5pt,
        xlabel = Goals,
        axis x line=bottom,
        axis y line=left,
        enlarge y limits=0.01, 
        xticklabels={26,27,28,29,30,31,32,33,34,35,36}
    ]
    \addplot table[x=shaphome,y expr=\coordindex, col sep=comma] {figures/data/ForcePlotMetz.csv};
    \draw[thick] (0,0) -- (0,27);
    \node[black!50] at (0.55,28) {\footnotesize Prediction: 32 goals};
    \end{axis}
\end{tikzpicture}}
	\caption{Force plot of predicted goals (from CatBoost with SEL) for Metz Handball for the game played on May 17\textsuperscript{th} against Chambray Handball}
	\label{fig:force_plot}
\end{figure}
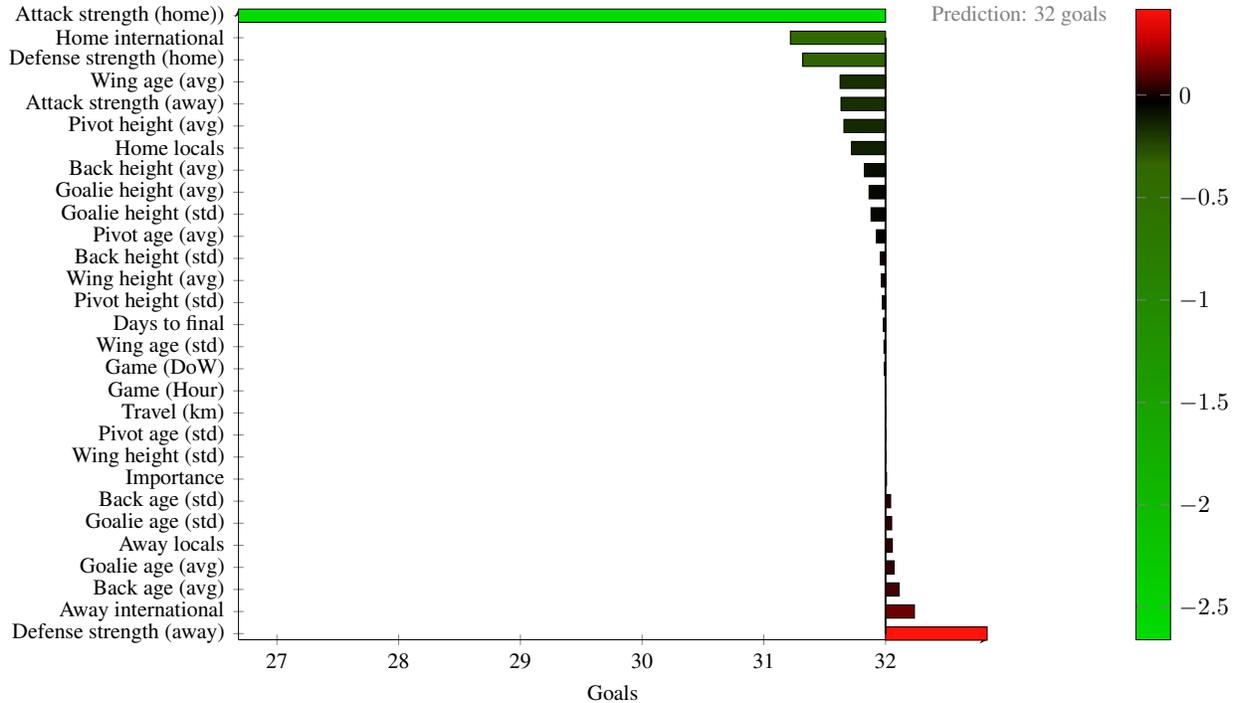

To that end, we analyze the last game of the season 2022/2023 played at home for Metz Handball.
This game was played on May 17\textsuperscript{th}, 2023 against Chambray Touraine Handball as part of the "Ligue Butagaz Energie" (LBE), the French female 1\textsuperscript{st} division championship.
We choose this game as it is not contained in the training set, the predictions were even made and explanations generated before the game\footnote{Predictions and explanations generated on May 16\textsuperscript{th}}.
Furthermore, the authors attended the game, which can help understand the generated explanations with concrete elements that happened during the actual game.
The model predicted a final score of 32-24 in favor of Metz Handball and the actual game saw Metz winning 30-26 over Chambray Touraine Handball.
We present in Figure~\ref{fig:force_plot} the explanations generated from the CatBoost model with SEL features for the selected game.

We can read Figure~\ref{fig:force_plot} as features close to the top of the plot contribute positively to increasing the number of goals Metz could score during the game.
On the other hand, features  at the bottom of the plot contribute negatively to goals scored by Metz, i.e. stand for the defense of Chambray.
Therefore, in line with our conclusions from the feature importance plot from Figure~\ref{fig:feat_imp}, the attack strength for the home team (Metz) is the main contributor for the final score.
The defense from Chambray, however, contributes to lower the total score, without which the outcome would be much worse for the team.
Other features such as the experience from international or wing players positively contributed to the victory of Metz.
An additional factor is the number of days until the final (end of season).
Although the model could not be aware that Metz, playing their last game of the season at home, was about to receive the trophy and celebrate the title of champion, the few days left until the end of season contributed to the motivation of players.

The model still remains based on statistical facts from past events, the explanations and predictions derived from the ML tool still echoed on concrete sports events during the actual match.

\section{Discussion}\label{sec:discussion}

We showed that the proposed ML solution achieves a high predictive performance and explanations generated with relevant explainability frameworks allow a translation of our analytical findings into concrete sports events.
In this section, we argue that this tool can be used by sports professionals such as team managers to prepare for upcoming games.
We also open the discussion for future work on extending  team strengths to player abilities  as additional SEL covariates.

\subsection{An analytical tool for coaches}

While state-of-the-art Machine Learning models for sports predictions (e.g. football, basketball) usually plateau around 75\% accuracy to predict the outcome of a match \citep{huang_neural_2010, lampis_predictions_2023}, our proposed solution for handball matches achieves above 80\% accuracy.
Coupled with explainability capabilities, our approach can translate statistical predictions into real facts happening during a match.
Although no model can guarantee that the result of a game will be as predicted, the model can identify statistical facts that can explain parts of the outcome.
Such patterns can therefore be used by team coaches in view of a competition.

Knowing the prediction of a game together with the potential main contributors to this outcome can help prepare a game and improve the team's strategy.
As we illustrated in Section~\ref{sec:model_explain_local}, local explainability can reveal where a team is expected to excel or struggle during the match.
We observed in Table~\ref{fig:force_plot} that a team can have an advantage with the experience of their wing players or goalkeepers and struggle due to the defense strength of the opposing team.
Therefore, a coach can use these pieces of information to ensure the team can accentuate on predicted strengths and work on removing their weak points.


\subsection{From team strengths estimation to player abilities}

As presented in Section~\ref{sec:sel_feat}, the structure of handball games suggests the use of the Conway-Maxwell-Poisson distribution from which we can derive a formula to estimate the attack and defense strengths of a team.
We showed the importance of this feature to the predictive performance of a model, and \cite{felice_ranking_2023} illustrated the relevance of such metrics to derive the ranking of clubs.
This methodology can be adapted to other settings such as the estimation of individual player abilities.
Although the publicly available data does not allow  extracting a long history of player statistics over multiple seasons, having access to such data could lead to similar research.

\section{Conclusions}\label{sec:conclusion}

In this paper, we showed how we can construct a highly accurate predictive model for handball games.
While data preparation and feature engineering are often under-explored in the literature \citep{zheng_feature_2018, felice_statistically_2023}, our results highlight their importance on the model's performance.
This encourages us to focus, in future works, on the preparation of even more meaningful features to capture more signals and further improve the model's performance.
The models presented in this paper are trained and evaluated on female championships but this work can {easily} be extended to male's championships as well as international competitions.
In view of the upcoming Olympic Games in Paris in 2024, the presented solution can also target national teams' coaches to prepare for this worldwide event by means of analytical tools powered by accurate Machine Learning models.

\bibliography{references2}

\begin{thebibliography}{}

\bibitem[Akyüz et~al., 2019]{akyuz_skeletal_2019}
Akyüz, B., Avşar, P.~A., Bilge, M., Deliceoğlu, G., and Korkusuz, F. (2019).
\newblock Skeletal muscle fatigue does not affect shooting accuracy of handball
  players.
\newblock {\em Isokinetics and Exercise Science}, 27(4):253--259.

\bibitem[Breiman, 2001]{breiman_random_2001}
Breiman, L. (2001).
\newblock Random {Forests}.
\newblock {\em Machine Learning}, 45(1):5--32.

\bibitem[Brier, 1950]{brier_verification_1950}
Brier, G.~W. (1950).
\newblock Verification of {Forecasts} {Expressed} in {Terms} of {Probability}.
\newblock {\em Monthly Weather Review}, 78(1):1--3.
\newblock Publisher: American Meteorological Society Section: Monthly Weather
  Review.

\bibitem[Cai et~al., 2019]{cai_hybrid_2019}
Cai, W., Yu, D., Wu, Z., Du, X., and Zhou, T. (2019).
\newblock A hybrid ensemble learning framework for basketball outcomes
  prediction.
\newblock {\em Physica A: Statistical Mechanics and its Applications},
  528:121461.

\bibitem[Camacho-Cardenosa et~al., 2018]{camacho-cardenosa_anthropometric_2018}
Camacho-Cardenosa, A., Camacho-Cardenosa, M., González-Custodio, A.,
  Martínez-Guardado, I., Timón, R., Olcina, G., and Brazo-Sayavera, J.
  (2018).
\newblock Anthropometric and {Physical} {Performance} of {Youth} {Handball}
  {Players}: {The} {Role} of the {Relative} {Age}.
\newblock {\em Sports}, 6(2):47.

\bibitem[Chen and Guestrin, 2016]{chen_xgboost_2016}
Chen, T. and Guestrin, C. (2016).
\newblock {XGBoost}: {A} {Scalable} {Tree} {Boosting} {System}.
\newblock In {\em Proceedings of the 22nd {ACM} {SIGKDD} {International}
  {Conference} on {Knowledge} {Discovery} and {Data} {Mining}}, {KDD} '16,
  pages 785--794, New York, NY, USA. Association for Computing Machinery.

\bibitem[Du et~al., 2019]{du_techniques_2019}
Du, M., Liu, N., and Hu, X. (2019).
\newblock Techniques for interpretable machine learning.
\newblock {\em Communications of the ACM}, 63(1):68--77.

\bibitem[Felice, 2023]{felice_ranking_2023}
Felice, F. (2023).
\newblock Ranking {Handball} {Teams} from {Statistical} {Strength}
  {Estimation}.
\newblock arXiv:2307.06754 [stat].

\bibitem[Felice et~al., 2023]{felice_statistically_2023}
Felice, F., Ley, C., Groll, A., and Bordas, S. (2023).
\newblock Statistically {Enhanced} {Learning}: a feature engineering framework
  to boost (any) learning algorithms.
\newblock arXiv:2306.17006 [stat].

\bibitem[Fonseca et~al., 2019]{fonseca_relative_2019}
Fonseca, F.~S., Figueiredo, L.~S., Gantois, P., De~Lima-Junior, D., and Fortes,
  L.~S. (2019).
\newblock Relative {Age} {Effect} is {Modulated} by {Playing} {Position} but is
  {Not} {Related} to {Competitive} {Success} in {Elite} {Under}-19 {Handball}
  {Athletes}.
\newblock {\em Sports}, 7(4):91.

\bibitem[Grabara, 2018]{grabara_posture_2018}
Grabara, M. (2018).
\newblock The posture of adolescent male handball players: {A} two-year study.
\newblock {\em Journal of Back and Musculoskeletal Rehabilitation},
  31(1):183--189.

\bibitem[Groll et~al., 2020]{groll_prediction_2020}
Groll, A., Heiner, J., Schauberger, G., and Uhrmeister, J. (2020).
\newblock Prediction of the 2019 {IHF} {World} {Men}’s {Handball}
  {Championship} – {A} sparse {Gaussian} approximation model.
\newblock {\em Journal of Sports Analytics}, 6(3):187--197.
\newblock Publisher: IOS Press.

\bibitem[Groll et~al., 2019]{groll_hybrid_2019}
Groll, A., Ley, C., Schauberger, G., and Van~Eetvelde, H. (2019).
\newblock A hybrid random forest to predict soccer matches in international
  tournaments.
\newblock {\em Journal of Quantitative Analysis in Sports}, 15(4):271--287.

\bibitem[Hahn et~al., 2013]{hahn_fascination_2013}
Hahn, R., Glock, R., and Birkefeld, F. (2013).
\newblock {\em Fascination for {Thousands} of {Years} - {Handball} {History}
  and {Stories}}.
\newblock International Handball Federation.

\bibitem[Hamon et~al., 2020]{hamon_robustness_2020}
Hamon, R., Junklewitz, H., and Sanchez, M. J.~I. (2020).
\newblock Robustness and {Explainability} of {Artificial} {Intelligence}.
\newblock ISBN: 9789276146605 ISSN: 1831-9424.

\bibitem[Huang and Chang, 2010]{huang_neural_2010}
Huang, K.-Y. and Chang, W.-L. (2010).
\newblock A neural network method for prediction of 2006 {World} {Cup}
  {Football} {Game}.
\newblock In {\em The 2010 {International} {Joint} {Conference} on {Neural}
  {Networks} ({IJCNN})}, pages 1--8, Barcelona, Spain. IEEE.

\bibitem[Lampis et~al., 2023]{lampis_predictions_2023}
Lampis, T., Ioannis, N., Vasilios, V., and Stavrianna, D. (2023).
\newblock Predictions of european basketball match results with machine
  learning algorithms.
\newblock {\em Journal of Sports Analytics}, pages 1--20.

\bibitem[Lundberg and Lee, 2017]{lundberg_unified_2017}
Lundberg, S. and Lee, S.-I. (2017).
\newblock A {Unified} {Approach} to {Interpreting} {Model} {Predictions}.
\newblock Number: arXiv:1705.07874 arXiv:1705.07874 [cs, stat].

\bibitem[Lundberg et~al., 2020]{lundberg_local_2020}
Lundberg, S.~M., Erion, G., Chen, H., DeGrave, A., Prutkin, J.~M., Nair, B.,
  Katz, R., Himmelfarb, J., Bansal, N., and Lee, S.-I. (2020).
\newblock From local explanations to global understanding with explainable {AI}
  for trees.
\newblock {\em Nature Machine Intelligence}, 2(1):56--67.
\newblock Number: 1 Publisher: Nature Publishing Group.

\bibitem[Madsen et~al., 2019]{madsen_activity_2019}
Madsen, M., Ermidis, G., Rago, V., Surrow, K., Vigh-Larsen, J.~F., Randers,
  M.~B., Krustrup, P., and Larsen, M.~N. (2019).
\newblock Activity {Profile}, {Heart} {Rate}, {Technical} {Involvement}, and
  {Perceived} {Intensity} and {Fun} in {U13} {Male} and {Female} {Team}
  {Handball} {Players}: {Effect} of {Game} {Format}.
\newblock {\em Sports}, 7(4):90.

\bibitem[McCabe and Trevathan, 2008]{mccabe_artificial_2008}
McCabe, A. and Trevathan, J. (2008).
\newblock Artificial {Intelligence} in {Sports} {Prediction}.
\newblock In {\em Fifth {International} {Conference} on {Information}
  {Technology}: {New} {Generations} (itng 2008)}, pages 1194--1197, Las Vegas,
  NV, USA. IEEE.

\bibitem[Miljkovic et~al., 2010]{miljkovic_use_2010}
Miljkovic, D., Gajic, L., Kovacevic, A., and Konjovic, Z. (2010).
\newblock The use of data mining for basketball matches outcomes prediction.
\newblock In {\em {IEEE} 8th {International} {Symposium} on {Intelligent}
  {Systems} and {Informatics}}, pages 309--312, Subotica, Serbia. IEEE.

\bibitem[Olympics, 2023]{olympics_history_2023}
Olympics (2023).
\newblock History of handball.
\newblock Accessed: 2023-05-13.

\bibitem[Pic, 2018]{pic_performance_2018}
Pic, M. (2018).
\newblock Performance and {Home} {Advantage} in {Handball}.
\newblock {\em Journal of Human Kinetics}, 63:61--71.

\bibitem[Prokhorenkova et~al., 2018]{prokhorenkova_catboost_2018}
Prokhorenkova, L., Gusev, G., Vorobev, A., Dorogush, A.~V., and Gulin, A.
  (2018).
\newblock {CatBoost}: unbiased boosting with categorical features.
\newblock In {\em Proceedings of the 32nd {International} {Conference} on
  {Neural} {Information} {Processing} {Systems}}, {NIPS}'18, pages 6639--6649,
  Red Hook, NY, USA. Curran Associates Inc.

\bibitem[Ribeiro et~al., 2016]{ribeiro_why_2016}
Ribeiro, M.~T., Singh, S., and Guestrin, C. (2016).
\newblock "{Why} {Should} {I} {Trust} {You}?": {Explaining} the {Predictions}
  of {Any} {Classifier}.
\newblock Number: arXiv:1602.04938 arXiv:1602.04938 [cs, stat].

\bibitem[Rodriguez-Ruiz et~al., 2011]{rodriguez-ruiz_study_2011}
Rodriguez-Ruiz, D., Quiroga, M.~E., Miralles, J.~A., Sarmiento, S., De~Saá,
  Y., and García-Manso, J.~M. (2011).
\newblock Study of the {Technical} and {Tactical} {Variables} {Determining}
  {Set} {Win} or {Loss} in {Top}-{Level} {European} {Men}'s {Volleyball}.
\newblock {\em Journal of Quantitative Analysis in Sports}, 7(1).

\bibitem[Rosenblatt, 1958]{rosenblatt_perceptron_1958}
Rosenblatt, F. (1958).
\newblock The perceptron: {A} probabilistic model for information storage and
  organization in the brain.
\newblock {\em Psychological Review}, 65(6):386--408.

\bibitem[Saavedra, 2018]{saavedra_handball_2018}
Saavedra, J.~M. (2018).
\newblock Handball {Research}: {State} of the {Art}.
\newblock {\em Journal of Human Kinetics}, 63(1):5--8.

\bibitem[Seil et~al., 1998]{seil_sports_1998}
Seil, R., Rupp, S., Tempelhof, S., and Kohn, D. (1998).
\newblock Sports {Injuries} in {Team} {Handball}.
\newblock {\em The American Journal of Sports Medicine}, 26(5):681--687.

\bibitem[Sellers, 2022]{sellers_conway-maxwell-poisson_2022}
Sellers, K.~F. (2022).
\newblock {\em The {Conway}-{Maxwell}-{Poisson} distribution}.
\newblock Institute of {Mathematical} {Statistics} monographs. Cambridge
  University Press, Cambridge, United Kingdom ; New York, NY, USA, first
  edition edition.

\bibitem[Sovrano et~al., 2022]{sovrano_metrics_2022}
Sovrano, F., Sapienza, S., Palmirani, M., and Vitali, F. (2022).
\newblock Metrics, {Explainability} and the {European} {AI} {Act} {Proposal}.
\newblock {\em J}, 5(1):126--138.

\bibitem[Wagner et~al., 2014]{wagner_individual_2014}
Wagner, H., Finkenzeller, T., Würth, S., and von Duvillard, S.~P. (2014).
\newblock Individual and team performance in team-handball: a review.
\newblock {\em Journal of Sports Science \& Medicine}, 13(4):808--816.

\bibitem[Zebari et~al., 2021]{zebari_predicting_2021}
Zebari, G.~M., Zeebaree, S., M.Sadeeq, M., and Zebari, R. (2021).
\newblock Predicting {Football} {Outcomes} by {Using} {Poisson} {Model}:
  {Applied} to {Spanish} {Primera} {División}.
\newblock {\em Journal of Applied Science and Technology Trends},
  2(04):105--112.
\newblock Number: 04.

\bibitem[Zheng and Casari, 2018]{zheng_feature_2018}
Zheng, A. and Casari, A. (2018).
\newblock {\em Feature engineering for machine learning: principles and
  techniques for data scientists}.
\newblock "O'Reilly Media, Inc.".

\end{thebibliography}

\newpage


\appendix
\section{Appendix}
\subsection[\appendixname~\thesubsection]{List of features}\label{apx:features}


\begin{table}[H] 
\caption{Complete list of features with data type and description.\label{tab:features}}
\newcolumntype{C}{>{\centering\arraybackslash}X}
\begin{tabularx}{\textwidth}{lcC}
\toprule
\textbf{Attribute}	& \textbf{Data type}	  & \textbf{Description}\\
\midrule
\texttt{game\_dow}					& Integer & Day of the week of the game.\\
\texttt{game\_hour}					& Integer & Hour of the start time of the game.\\
\texttt{importance}					& Integer & Importance of the competition from the lowest (friendly games with value 5) to the highest importance (Olympic Games with value 1)\\
\texttt{days\_to\_final} 			& Integer & Number of days until the competition's final or end of season.\\
\texttt{away\_travel\_km} 			& Float	  & Distance in kilometers (as the crow flies) between the away and home teams' locations.\\
\texttt{home\_international}        & Float   & Share of international players (for ongoing season) in home team.\\
\texttt{away\_international}        & Float   & Share of international players (for ongoing season) in away team.\\
\texttt{home\_locals}               & Float   & Share of home players with the same nationality as the club (to reflect barriers from language.\\
\texttt{away\_locals}               & Float   & Share of away players with the same nationality as the club (to reflect barriers from language.\\
\texttt{diff\_wing\_height\_avg}	& Float	& Difference of the average height of wing players between home and away teams.\\
\texttt{diff\_back\_height\_avg}	& Float & Difference of the average height of back (left, center and right) players between home and away teams.\\
\texttt{diff\_pivot\_height\_avg}	& Float & Difference of the average height of line (aka. pivot) players between home and away teams.\\
\texttt{diff\_gk\_height\_avg}		& Float & Difference of the average height of goalkeepers between home and away teams.\\
\texttt{diff\_wing\_height\_std}	& Float	& Difference of the standard deviation of heights  of wing players between home and away teams.\\
\texttt{diff\_back\_height\_std}	& Float & Difference of the   standard deviation  of heights  of back (left, center and right) players between home and away teams.\\
\texttt{diff\_pivot\_height\_std}	& Float & Difference of the standard deviation of heights of line (aka. pivot) players between home and away teams.\\
\texttt{diff\_gk\_height\_std}		& Float & Difference of the standard deviation of heights  of goalkeepers between home and away teams.\\
\texttt{diff\_wing\_weight\_avg}	& Float	& Difference of the average weight of wing players between home and away teams.\\
\texttt{diff\_back\_weight\_avg}	& Float & Difference of the average weight of back (left, center and right) players between home and away teams.\\
\texttt{diff\_pivot\_weight\_avg}	& Float & Difference of the average weight of line (aka. pivot) players between home and away teams.\\
\texttt{diff\_gk\_weight\_avg}		& Float & Difference of the average weight of goalkeepers between home and away teams.\\
\texttt{diff\_wing\_weight\_std}	& Float	& Difference of the standard deviation of weights  of wing players between home and away teams.\\
\texttt{diff\_back\_weight\_std}	& Float & Difference of the  standard deviation of weights  of back (left, center and right) players between home and away teams.\\
\texttt{diff\_pivot\_weight\_std}	& Float & Difference of the standard deviation of weights  of line (aka. pivot) players between home and away teams.\\
\texttt{diff\_gk\_weight\_std}		& Float & Difference of the standard deviation of weights  of goalkeepers between home and away teams.\\
\texttt{diff\_wing\_age\_avg}		& Float	& Difference of the average age of wing players between home and away teams.\\
\texttt{diff\_back\_age\_avg}		& Float & Difference of the average age of back (left, center and right) players between home and away teams.\\
\texttt{diff\_pivot\_age\_avg}		& Float & Difference of the average age of line (aka. pivot) players between home and away teams.\\
\bottomrule
\end{tabularx}
\end{table}

\begin{table}[H] 
\newcolumntype{C}{>{\centering\arraybackslash}X}
\begin{tabularx}{\textwidth}{lcC}
\toprule
\textbf{Attribute}	& \textbf{Data type}	  & \textbf{Description}\\
\midrule
\texttt{diff\_gk\_age\_avg}			& Float & Difference of the average age of goalkeepers between home and away teams.\\
\texttt{diff\_wing\_age\_std}		& Float	& Difference of the standard deviation of ages  of wing players between home and away teams.\\
\texttt{diff\_back\_age\_std}		& Float & Difference of the standard deviation of ages  of back (left, center and right) players between home and away teams.\\
\texttt{diff\_pivot\_age\_std}		& Float & Difference of the standard deviation of ages  of line (aka. pivot) players between home and away teams.\\
\texttt{diff\_gk\_age\_std}			& Float & Difference of the standard deviation of ages  of goalkeepers between home and away teams.\\
\texttt{attack\_strength\_home}		& Float & Attack strength estimated via SEL for home team.\\
\texttt{defense\_strength\_home}	& Float & Defense strength estimated via SEL for home team.\\
\texttt{attack\_strength\_away}		& Float & Attack strength estimated via SEL for away team.\\
\texttt{defense\_strength\_away}	& Float & Defense strength estimated via SEL for away team.\\
\bottomrule
\end{tabularx}
\end{table}

\end{document}